# Hybridization of Otsu Method and Median Filter for Color Image Segmentation

Firas Ajil Jassim, Fawzi H. Altaani

*Abstract—In this article a novel algorithm for color image segmentation has been developed. The proposed algorithm based on combining two existing methods in such a novel way to obtain a significant method to partition the color image into significant regions. On the first phase, the traditional Otsu method for gray channel image segmentation were applied for each of the R,G, and B channels separately to determine the suitable automatic threshold for each channel. After that, the new modified channels are integrated again to formulate a new color image. The resulted image suffers from some kind of distortion. To get rid of this distortion, the second phase is arise which is the median filter to smooth the image and increase the segmented regions. This process looks very significant by the ocular eye. Experimental results were presented on a variety of test images to support the proposed algorithm.*

## I. INTRODUCTION

One of the most important problems in color image analysis is that of segmentation. The fundamental idea in color image segmentation is to consider color uniformity as a relevant criterion to partition an image into significant regions [9]. The first task of any image processing tasks is usually Segmentation. All subsequent tasks depend surely on the nature of segmentation. For this reason, a worthwhile attention is taken to improve the quality of segmentation [13]. Most care to image segmentation has been concentrated on gray level images. A common problem in the segmentation of gray scale images occurs when an image has a varying gray level background. One of these problems is the progressively varying shadows, or when the image contains wide range of gray levels. This is an inveterate intensity problem in gray scale images. According to [11], the detection procedure for the human in a high details image can be done for one or two dozen intensity levels at any point due to luminosity accommodation. On the other hand, thousands of color shadows and intensities could be distinguished by the human eye can.

Nowadays, there is no robust mathematical theory of image segmentation. Consequently, no solitary typical method of image segmentation has originated. Therefore, there are a variety of solo methods that have been some what popular according to its success [14]. Generally, most of the segmentation techniques for gray scale images such as histogram thresholding, edge detection, feature clustering, fuzzy methods, region based methods, and neural networks have been extended for color image segmentation by using RGB color space system or other color space like CYM, HSI, etc. Anyway, recently, there is a shortage in the comprehensive surveys on color image segmentation [7]. Color images can convey more information than gray scale images [4]. Color image segmentation is a method of mining one or more unified regions that are homogenous. This may be obtained from region's spectral elements [15]. Recently, there are a large number of color image segmentation techniques. They can be classified into four general categories: pixel-based, edge-based, region-based, and model-based techniques. Actually, the basic behavior of these techniques can be divided into three concepts. The first concept is the similarity concept like edge-based techniques. Alternatively, the second concept is based on the discontinuity of pixel values like pixel-based and region-based techniques. Finally, a complete different approach is the third concept which is based on a statistical approach like Model-based techniques. In the third concept, segmentation is implemented as an optimization problem [11].

The organization of this paper is as follows: In the next section a clement debate about traditional Otsu method has been introduced. In section (III), a passing through median filter was quietly presented. The proposed technique for color image segmentation which is the basic contribution of this paper has been discussed in section (IV). In section (V), the practical implementation with experimental results was introduced and supported with ocular evidences to assist the proposed technique. Finally, the conclusion section may be seen in section (VI) that contains the figuring out results.

## II. OTSU METHOD PRELIMINARIES

Otsu method is one of the oldest methods in image segmentation [12]. It is treated as a statistical method according to its probabilistic implementation. It must be mentioned that the Otsu method is one of the best automatic thresholding methods [3]. The basic principle in Otsu method is to split the image into two classes which are the objects and the background. The automatic threshold is obtained through finding the maximum variance between the two classes [16]. Practically speaking, let $I=[1,L]$ is the range of grayscale levels of image $f(x,y)$ and $p_i$ is the probability of each level. The number of pixels with gray level i is denoted $f_i$, giving a probability of gray level i in an image as:

$$p_i = \frac{f_i}{N} \qquad (1)$$

Then, the automatic threshold t that divides the range into two classes which are $C_0=[1, …, t]$ and $C_1=[t+1, …, L]$ [10]. Then, the gray level probability distributions for the two classes are:

$$C_1 \rightarrow \left[ \frac{p_1}{\sum_{i=1}^{t} p_i}, \frac{p_2}{\sum_{i=1}^{t} p_i}, ..., \frac{p_t}{\sum_{i=1}^{t} p_i} \right] \quad (2)$$

$$C_2 \rightarrow \left[ \frac{p_{t+1}}{\sum_{i=t+1}^{L} p_i}, \frac{p_{t+2}}{\sum_{i=t+1}^{L} p_i}, ..., \frac{p_L}{\sum_{i=t+1}^{L} p_i} \right] \quad (3)$$

Also, the means for classes $C_1$ and $C_2$ are:

$$\mu_1 = \frac{\sum_{i=1}^{t} i.p_i}{\sum_{i=1}^{t} p_i} \quad (4)$$

$$\mu_2 = \frac{\sum_{i=t+1}^{L} i.p_i}{\sum_{i=t+1}^{L} p_i} \quad (5)$$

Let $\mu_T$ be the overall mean of the whole image. Obviously, by summing the parts, it is easy to show that:

$$\beta_1 \mu_1 + \beta_2 \mu_2 = \mu_T$$

where

$$\beta_1 = \sum_{i=1}^{t} p_i \text{ and } \beta_2 = \sum_{i=t+1}^{L} p_i$$

From statistics, it is clear that the total sum of the probabilities is always equal to one.

$$\beta_1 + \beta_2 = 1$$

Finally, Otsu defined the between-class variance two classes $C_1$ and $C_2$ as:

$$\sigma^2 = \beta_1 (\mu_1 - \mu_T)^2 + \beta_2 (\mu_2 - \mu_T)^2 \quad (6)$$

In case of bi-level thresholding, Otsu proved that the optimal threshold $t'$ is the value that maximize the between-class variance $\sigma^2$ as,

$$t' = \max\{\sigma^2(t)\}, \ 1 \leq t < L \quad (7)$$

## III. MEDIAN FILTER

Median filtering is one of the best methods that used to suppress Salt & Peppers noise which is called impulse noise. Median filter has demonstrated an effective technique to suppress impulse noise while preserving signal changes [6]. The classical median filter replaces the central pixel in a (k×k) window with the median of the pixels inside that window. The median is the central location of the pixels after arranging these pixels in ascending order. As an example, in a 5×5 window we pick up the thirteenth value in the ordered pixels after ascending arrangement [5]. Median filter aims to change noisy pixels in such a way to be look like its nearby neighbors [1]. Median filtering removes noise without blurring edges when the window size is reasonable (small), but it also makes rooftop edges tabulate [8]. The fundamental consequence of median filtering is that pixels with noisy values are compelled to have nearly similar value like their surrounding neighbors.

Actually, median filter is a non-linear filter that can be used to smooth images [2]. Furthermore, one disadvantage of median filter is that high blurring in the image when large window size is implemented [13]. From this point, the main contribution is this paper has come arise. Therefore, when applying median filters as a second phase after the traditional Otsu method, the resulted segmented image is highly acceptable for the researchers in the field of color image segmentation.

## IV. PROPOSED SEGMENTATION METHOD

The proposed method starts with applying the standard Otsu method for automatic thresholding segmentation for each of the R, G, and B arrays in the digital image. The implementation of Otsu method was implemented separately, i.e. for each of the R, G, and B channels may be treated as a gray channel. Hence, the obtained segmented image for each channel was obtained by different threshold than the threshold for the other channels. The symbolic representation in this paper for the three thresholds for R,G,B are $T_R$, $T_G$, and $T_B$, respectively. Now, the new segmented image may be obtained by merging the thresholded R, G, and B channels together. Moreover, the median filtering technique could be applied to smooth the resulted image more. As stated in the previous section, the main contribution in this paper is that, when increasing the window size in the median filtering process, noticeable blurring results may be clear for the human eye. This pint of view treated as a disadvantage in the median filtering process, but from our point of view it is a benefit and could be useful in image segmentation. Hence, a suitable window size must be determined carefully is such a way that keeps the blurring amount in the safety side and away from the distortion and high blurring. The suitable window size that was obtained by the proposed algorithm was found to be (15×15). Actually, it must be mentioned that, when using the proposed window size (15×15) alone, i.e. without preceded by Otsu method, will produce ill consequences. Therefore, combining these two methods together (Otsu and median filter) will originate a novel method for color image segmentation.

According to figure (1a), it can be seen that the implementation of traditional Otsu method for each channel of the R,G,B channels will produce some kind of noisy regions. Therefore, to make these regions smoother, a median filter could be applied with k×k window size to get rid of these noisy regions. This process is very useful in object recognition, consequently, image segmentation. It must be mentioned that whenever there is an increase in the block size, there will be an increase in the smoothness process. Hence, seven types of block sizes have been applied which are: 3×3, 5×5, 7×7, 9×9, 11×11, 13×13, and 15×15. As a sample case, a 7×7 window size was presented in figure (1b) after Otsu method. The resulted image, figure (1b), is high segmented than the traditional Otsu method (figure 1a).

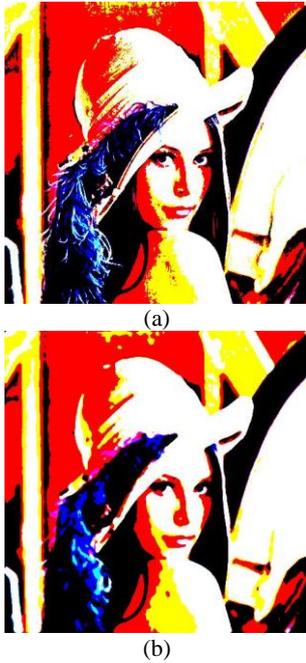

(a)

(b)

Fig. 1 (a) Original image (b) Traditional Otsu image for each of the R,G,B channels (c) Hybridization between Otsu method and 7×7 Median filter

## V. EXPERIMENTAL RESULTS

In this Section, details of the implementation and experimental results are presented concerning the previously proposed method. Several (512×512) test images have been used to implement the proposed method for color image segmentation, figure (2). The experimental results have been shown in figures (3), (4), (5), (6), and (7). Furthermore, a combination of window sizes have been applied which are (3×3, 5×5, 7×7, 9×9, 11×11, 13×13, and 15×15) in case of median filtering process. Obviously, from the shown results it is clear that there is a direct proportional relation between increasing window size and good clarity of segmentation results. But this increment in window size must be limited to avoid distortion in the reconstructed image at large window size. Hence, from the ocular results it is very clear that the convenient window size is (15×15) for all the test images that have been used.

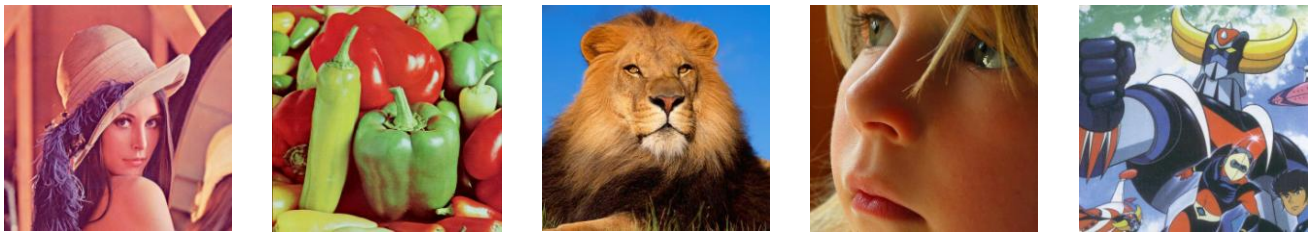

Fig. 2 Variety of (512×512) test images (Lena, Peppers, Lion, baby girl, and Grendizer)

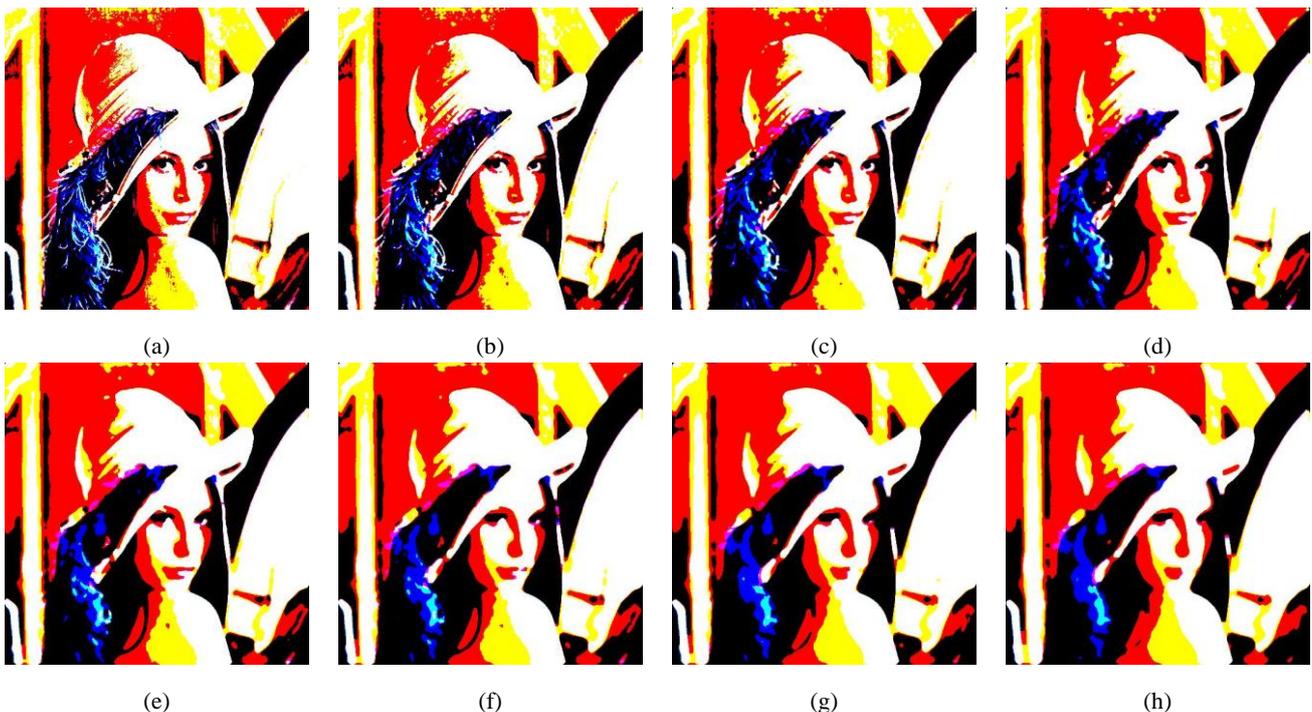

(a) (b) (c) (d)

(e) (f) (g) (h)

Fig. 3 (a) Otsu (b) Proposed 3×3 (c) Proposed 5×5 (d) Proposed 7×7 (e) Proposed 9×9 (f) Proposed 11×11 (g) Proposed 13×13 (h) Proposed 15×15

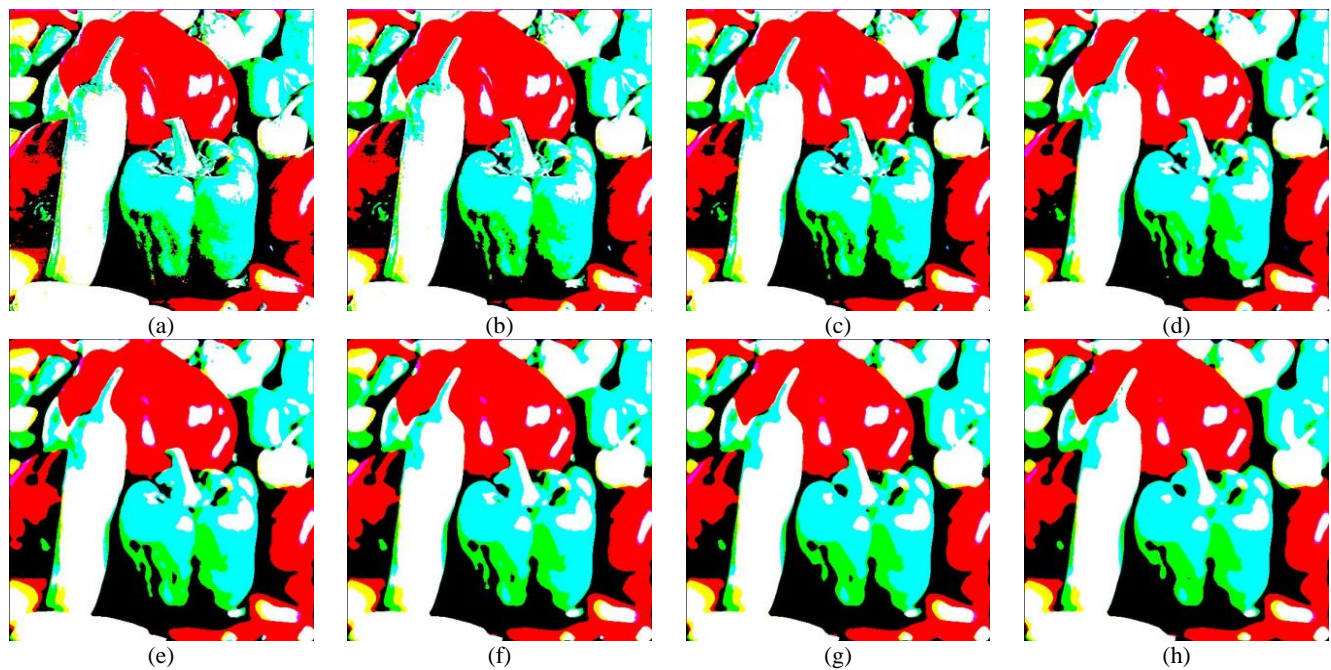

Fig. 4 (a) Otsu (b) Proposed 3×3 (c) Proposed 5×5 (d) Proposed 7×7 (e) Proposed 9×9 (f) Proposed 11×11 (g) Proposed 13×13 (h) Proposed 15×15

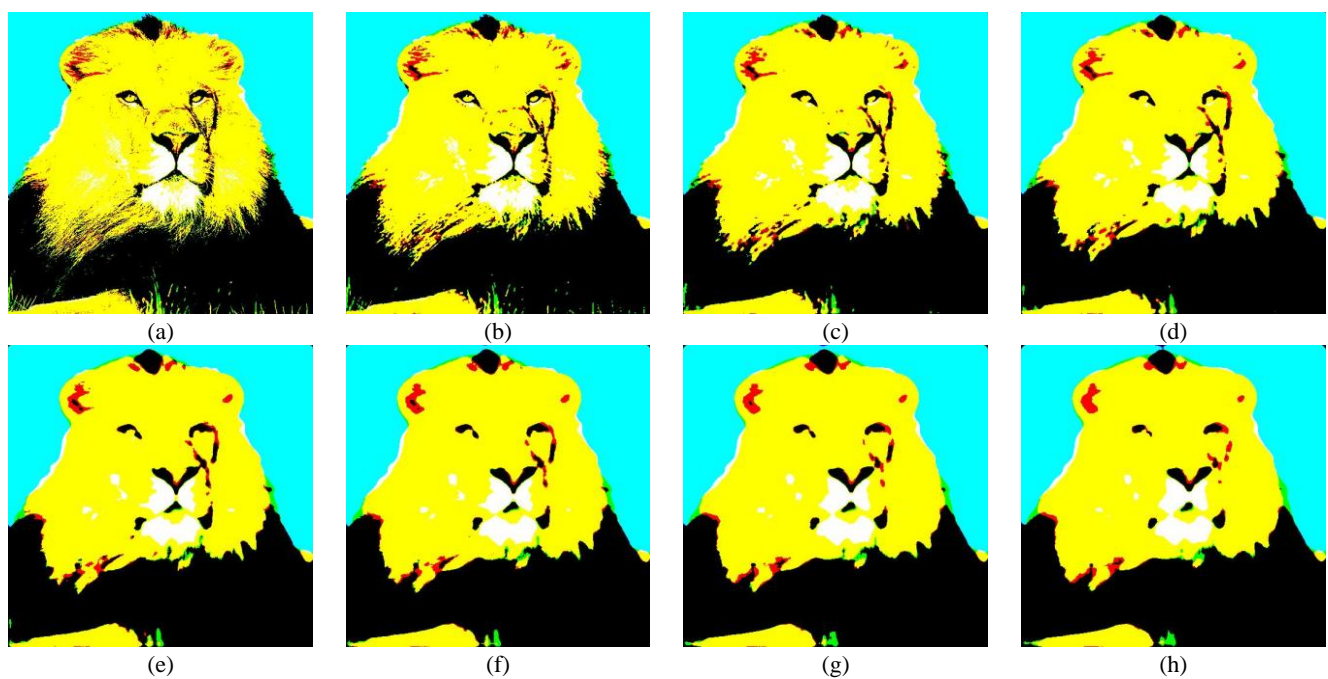

Fig. 5 (a) Otsu (b) Proposed 3×3 (c) Proposed 5×5 (d) Proposed 7×7 (e) Proposed 9×9 (f) Proposed 11×11 (g) Proposed 13×13 (h) Proposed 15×15

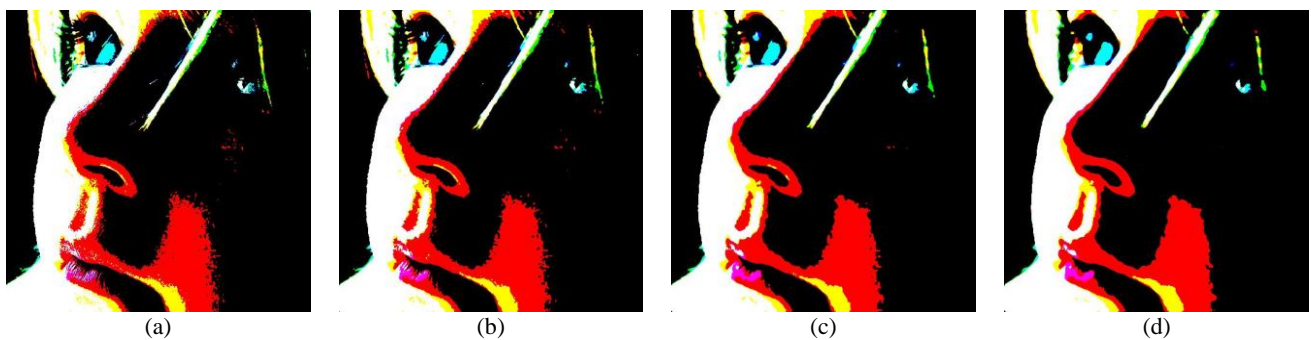

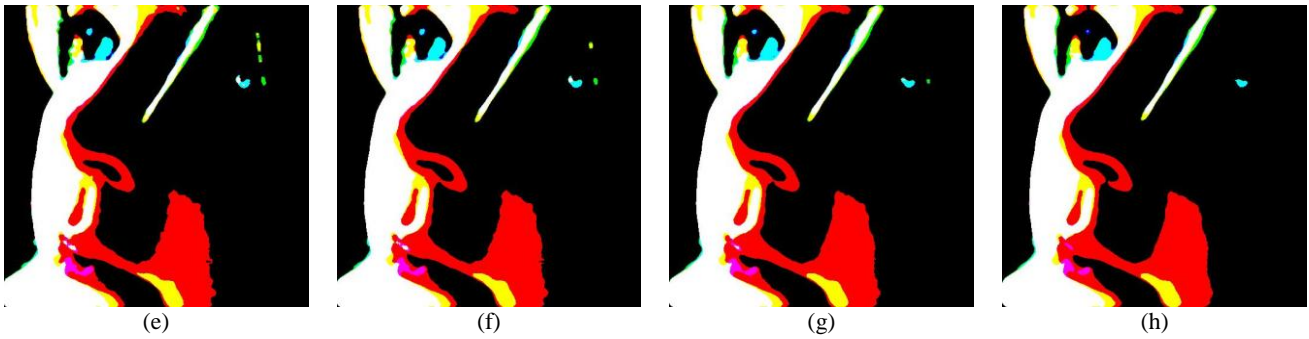

Fig. 6 (a) Otsu (b) Proposed 3×3 (c) Proposed 5×5 (d) Proposed 7×7 (e) Proposed 9×9 (f) Proposed 11×11 (g) Proposed 13×13 (h) Proposed 15×15

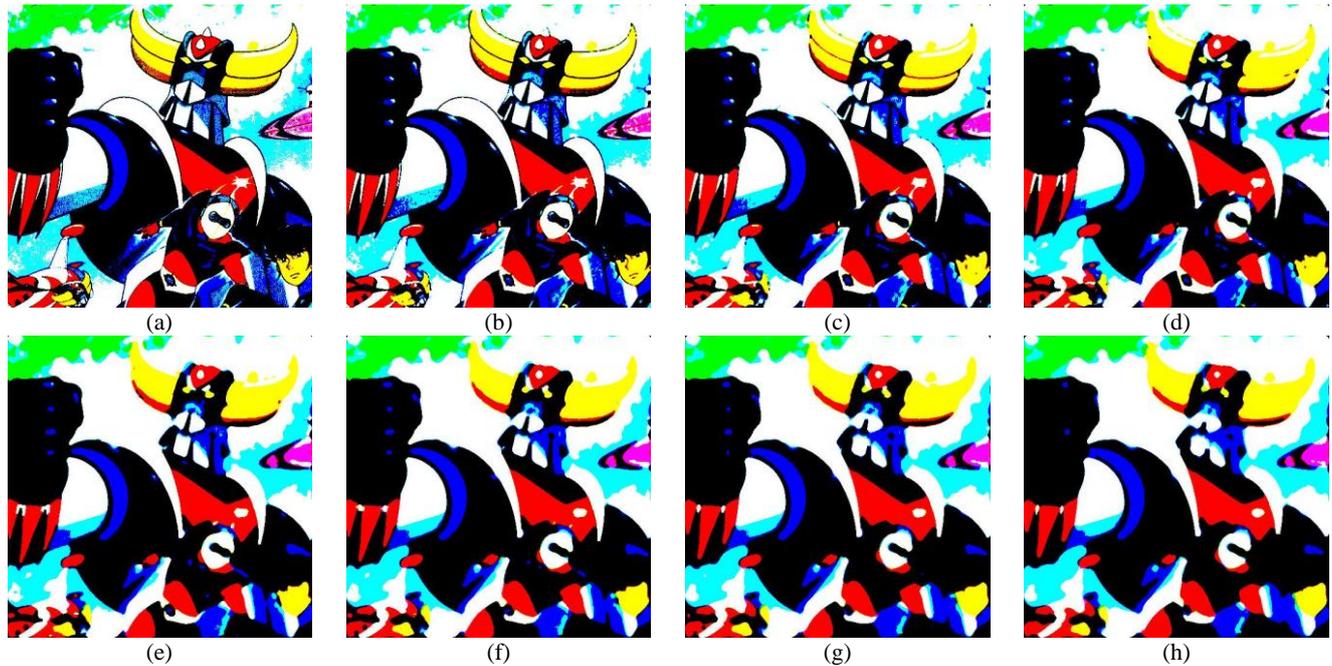

Fig. 7 (a) Otsu (b) Proposed 3×3 (c) Proposed 5×5 (d) Proposed 7×7 (e) Proposed 9×9 (f) Proposed 11×11 (g) Proposed 13×13 (h) Proposed 15×15

## I. COCLUSIONS

In this paper, a new approach for color image segmentation has been presented that is based on the hybridization between the classical Otsu method for gray level segmentation and Median filter. The implementation of Otsu method to the R,G,B channels alone will produce some kind of noise and to get rid of this noise a median filtering process was proposed. The implantation of median filter must be careful because it may cause some blurring in image when increasing the window size. The main conclusion comes here is that the increase in window size (k×k) that was implemented in filtering process will increase the interior homogeneity of the regions and objects inside the image. Hence, in this paper, a 15×15 windows size seems to be rational when applied to a variety of test images. Moreover, this method is too easy to implement concerning its simplicity and high rapidity. According to visual result in the previous section, and as a future work, the proposed technique is recommended in medical image processing, especially tumors detection procedure. Another question may be arise as a future work and that is what about implement the proposed technique on another color space like CMY or HIS or other.

**Firas A. Jassim** received the BS degree in Mathematics and Computer Applications from Al-Nahrain University, Baghdad, Iraq in 1997, and the MS degree in Mathematics and Computer Applications from Al-Nahrain University, Baghdad, Iraq in 1999 and the PhD degree in Computer Information Systems from the University of Banking and Financial Sciences, Amman, Jordan in 2012. His research interests are Image processing, image compression, image segmentation, image enhancement, image interpolation, and simulation.

**Fawzi Altaani** received the BS degree in Public Administration from Al-Yermouk University, Irbid, Jordan 1990, and Higher Diploma in Health Service Administration from University of Jordan, 1991 and the MS degree in Health Administration from Red Sea University, Sudan in 2004 and the PhD degree in Management Information Systems from the University of Banking and Financial Sciences, Amman, Jordan in 2010. His research interests are management information system and public administration, and Image processing.